\documentclass[journal]{IEEEtran}
\usepackage{algorithm}
\usepackage{algorithmic}
\pdfoutput=1
\makeatletter
\newcommand\fs@norules{\def\@fs@cfont{\bfseries}\let\@fs@capt\floatc@ruled
  \def\@fs@pre{}%
  \def\@fs@post{}%
  \def\@fs@mid{\kern3pt}%
  \let\@fs@iftopcapt\iftrue}
\makeatother
\floatstyle{norules}
\restylefloat{algorithm}
\usepackage{graphicx}

\ifCLASSINFOpdf
\else
\fi
\usepackage{subcaption}

\usepackage{comment}
\usepackage[activate={true,nocompatibility},final,tracking=true,kerning=true,spacing=true,factor=1100,stretch=10,shrink=10]{microtype}

\begin{document}
%
\title{Comparative Analysis of Extreme Verification Latency Learning Algorithms}
%
%
%

\author{Muhammad~Umer,
        and~Robi Polikar
\thanks{M. Umer and R. Polikar are with the Department
of Electrical and Computer Engineering, Rowan University, Glassboro,
NJ, 08028 USA.}
\thanks{}
\thanks{Manuscript received ; revised .}}

%
%

\markboth{Journal of --- ,~Vol.~, No.~, Month~20--}%
{Umer \MakeLowercase{\textit{et al.}}: Comparative Analysis of Extreme Verification Latency Learning Algorithms From Three Different Perspectives: Accuracy, Computational Complexity, and Parameter Sensitivity}
%



\maketitle

\begin{abstract}
One of the more challenging real-world problems in computational intelligence is to learn from non-stationary streaming data, also known as concept drift. Perhaps even a more challenging version of this scenario is when -- following a small set of initial labeled data -- the data stream consists of unlabeled data only. Such a scenario is typically referred to as learning in initially labeled nonstationary environment, or simply as extreme verification latency (EVL). Because of the very challenging nature of the problem, very few algorithms have been proposed in the literature up to date. This work is a very first effort to provide a review of some of the existing algorithms (important/prominent) in this field to the research community. More specifically, this paper is a comprehensive survey and comparative analysis of some of the EVL algorithms to point out the weaknesses and strengths of different approaches from three different perspectives: classification accuracy, computational complexity and parameter sensitivity using several synthetic and real world datasets.
\end{abstract}


%
\IEEEpeerreviewmaketitle

\section{Introduction}
\IEEEPARstart{T}{he} fundamental goal in machine learning is to learn from data. Most machine learning algorithm, regardless of the availability of labeled data, make a fundamental assumption that data are drawn from a fixed but unknown distribution. This assumption implies that test or field data come from the same distribution as the training data. In reality, this assumption simply does not hold in many real world problems that generate data whose underlying distributions change over time. Network intrusion, web usage and user interest analysis, natural language processing, speech and speaker identification, spam detection, anomaly detection, analysis of financial, climate, medical, energy demand, or pricing data, as well as the analysis of signals from autonomous robots and devices, brain signal analysis, and bio-informatics are just a few examples of the real world problems where underlying distributions may -- and typically do -- change over time.

In machine learning, the challenge of making decisions in a changing environment is referred to as non-stationary learning.  This is a challenging problem, because the classifier needs to adapt to a new concept in the changing environment, while retaining the previously acquired knowledge that is still relevant to ensure a stable learning environment, a phenomenon commonly referred to as the stability-plasticity dilemma in literature \cite{grossberg1988nonlinear}. The fixed distribution assumption, essentially requiring the data to be drawn independently from an identical distribution (also referred to as independent and identically distributed - i.i.d.) renders traditional learning algorithms that make this assumption ineffective at best, misleading and inaccurate at worst on non-stationary distribution problems. Concept drift techniques \cite{muhlbaier2007multiple,karnick2008learning, Elwell11,Ditzler13,Kolter07,Street01} and domain adaptation approaches \cite{Shimodaira00,Candela09} have been developed to tackle two related but different issues related to non-stationary distributions:  domain adaptation techniques are designed to handle mismatched training and test distribution over a single time-step, while concept drift approaches are designed to track the data distributions over a streaming setting. However, both approaches assume that there is (preferably ample) labeled training data, and the potential scarcity or the high cost of obtaining labeled data is a major obstacle faced by these approaches. 

In an effort to reduce the amount of required labeled data, semi supervised learning (SSL) and active learning (AL) approaches have been employed. SSL approaches, of course, also require labeled data at each time step \cite{Ditzler11}, albeit in smaller quantities. Active learning (AL) is another approach to combat the limited availability of labeled data \cite{Capo13}, where the learner actively chooses which data instances -- if labeled -- would provide the most benefit. The unavailability of labeled data, particularly in streaming applications, gives rise to another problem, commonly referred to as \textit{verification latency} in the literature \cite{Marrs10},  where labeled data are not available at every time step. More specifically, verification latency refers to the scenario where labels of the training data becoming available only certain or some unspecified amount of time later,  significantly complicating the learning process. The duration of the lag in obtaining labeled data may not be known a priori, and/or may vary with time. The extreme case of this phenomenon, aptly named as the \textit{extreme verification latency}, is perhaps the most challenging case of all machine learning problems: labels for the training data are never available - except perhaps those provided initially, yet the classification algorithm is asked to learn and track a drifting distribution with no access to labeled data. There are few algorithms proposed in the literature by different researchers to provide solution to the problem of extreme verification latency, however the extensive comparative analysis of these different algorithms is missing. Noting the importance of this problem in many real world applications, availability of such analysis to the community is indeed needed.

The primary goal of this work is to provide a comprehensive analysis of the extreme verification latency (EVL) learning algorithms from three different perspectives i.e. classification accuracy, computational complexity, and parameter sensitivity. To the best of our knowledge, this is the first comprehensive work in this regard. EVL, which is also referred as \textit{initially labeled non-stationary environment} (ILNSE) in \cite{Dyer14} is an extremely important but difficult problem, therefore commonly ignored by the researchers. This work is an effort to motivate the researchers to provide more realistic and practical solution to the problem under discussion. In particular, this work considers the following important algorithms proposed in the literature to work under EVL setting; i) \underline{A}rbitrary Sub-\underline{P}opulation \underline{T}racker (APT); \cite{Krempl11}, ii) \underline{COMP}acted \underline{O}bject \underline{S}ample \underline{E}xtraction (COMPOSE) \cite{Dyer14}; iii) \underline{S}tream \underline{C}lassification \underline{A}lgo\underline{r}ithm \underline{G}uided by \underline{C}lustering (SCARGC) \cite{Souza15_SCARGC}; iv) and \underline{M}icro-cluster for  \underline{C}lassification (MClassification) \cite{Souza15_MClassification}. 
\section{Extreme Verification Latency Learning Algorithms}


\subsection{Arbitrary Sub-Population Tracker Algorithm (APT)} 
APT algorithm is proposed by Krempl \cite{Krempl11} to handle extreme verification latency scenarios under specific scenarios. The main principle underlying APT algorithm is that each class in the data can be represented as a mixture of arbitrarily distributed sub-populations. The APT algorithm makes the following important assumptions
\begin{enumerate}
\item The underlying population of the feature space contains several sub-populations, each of which drifts (possibly) differently over time;
\item 	Initial labeled data are used to represent each sub-population of the feature space, where a sub-population is defined as a mode in the class-conditional distribution $p(y|x)$, with $p(y)$ representing the prior distribution of  the class labels, and $p(x)$ representing the marginal feature distribution; 
\item The drift is gradual and “systematic” that can be represented as a piecewise linear function;
\item The conditional posterior distribution remains fixed, i.e., a component’s class label cannot change
\item Co-variance of each component remains constant. 
\end{enumerate}
The learning strategy of APT is twofold; first, the optimal one-to-one assignment between labeled instances in time-step $t$ and unlabeled instances in time-step $t+1$ is determined using expectation maximization (EM) algorithm. The EM algorithm begins with the expectation step by predicting which instances are most likely to correspond to a given sub-population. During the maximization step, the algorithm determines which drift parameters maximize the expectation. Then, the classifier is updated to reflect the population parameters of the newly received data and drift parameter relating the previous time step to the current one. 

Establishing a one-to-one relationship while identifying drift requires an impractical assumption that the number of instances remains constant throughout all time steps. Krempl relaxes this assumption by establishing a relationship in a batch method - matching a random subset of exemplars to a subset of new observation until all new observations have been assigned a relationship to an exemplar. Krempl suggests a bootstrap method that can make the one-to-one assignments more robust, but at an additional computational cost. When the assumptions are satisfied, APT works very well. However, APT has two primary weaknesses: 1) some of its  assumptions often do not hold true, causing a decrease in performance, and 2) it is computationally very expensive \cite{Dyer14}.

The pseudocode for APT algorithm is given in Algorithm \ref{alg:apt}.

\begin{algorithm}[H]
 \caption{\textbf{Arbitrary Subpopulation Tracker (APT)}}
 \label{alg:apt}
 \begin{algorithmic}[1]
 \renewcommand{\algorithmicrequire}{\textbf{Inputs:}}
 \REQUIRE Initial labeled data $D_{init}$; A clustering algorithm with its own free parameters; a suitable bandwidth matrices calculation algorithm; a suitable expectation-maximization (EM) algorithm with its free parameters) 
\STATE Receive $M$ training examples form $D_{init} = $ $\lbrace x_i;y_i\rbrace$; $i = 1,...,M$ ; $x\in X ; y \in Y=\lbrace1,...,c\rbrace$;
\STATE Run clustering algorithm to partition the data into $K$ disjoint subsets and associate each cluster to one class among $c$ classes \label{clust_step};
\STATE Estimate the conditional feature distribution of the data;
\STATE Receive new unlabeled instances $U^t$ = $\left\{\textnormal{$x_u^t$ $\in$ $X$ , $u = {1,...,N}$}\right\}$ and assume $N = M$ to associate each new instance to one previous example;\label{unlabel}
\STATE Compute instance-to-exemplar correspondence  by maximizing the likelihood using EM algorithm;
\STATE Pass the cluster assignment from the example to their assigned instances to achieve instance-to-cluster assignment;
\STATE Pass the class of an example $x_i$ i.e. $y_i$ to the class of its assigned instance;
\STATE Go to step \ref{clust_step} and Repeat.
 \end{algorithmic}
 \end{algorithm}

\subsection{\textbf{COMP}acted \textbf{O}bject \textbf{S}ample \textbf{E}xtraction (COMPOSE)}
\subsubsection{COMPOSE.V1 (Original COMPOSE With $\alpha$-Shape Construction)}
The \textbf{COMP}acted \textbf{O}bject \textbf{S}ample \textbf{E}xtraction (COMPOSE) framework is introduced in \cite{Dyer14} to address the extreme verification latency problem in an extreme verification latency setting. The algorithm only makes an assumption of gradual/limited drift in the data, and consists of two important modules: semi-supervised learning algorithm (SSL) and the core-support extraction (CSE) module. It is an iterative procedure that uses an SSL algorithm to label the current unlabeled data using the initial labeled data. It then uses the core support extraction module to construct $\alpha$ shapes for each class and thus represent the current class conditional distribution, where $\alpha$-shape can be described as a generalization of the convex hull of the dataset, where the convex hull of a dataset $X \in {\rm I\!R}^d$ is the convex shape with minimum area that contains all of the observations in $X$, and can be described as the set of all possible convex combinations of the points in $X$, or 

\begin{equation}
{\lbrace \sum_{i=1}^{|X|}a_ix_i|(\forall i: a_i \geq 0) \space \wedge \sum_i a_i = 1 \rbrace }
\end{equation}
for all possible $a_i$.

The $\alpha$ shape is then compacted (shrunk), creating the core support region, and instances that fall inside this region are extracted as the core supports that represent the geometric center (core support region) of each class distribution. These now-labeled instances are used as the labeled information -- along with the incoming new unlabeled data -- to train the SSL algorithm during the next time step. This process is repeated every time there is a new batch of unlabeled data available. The pseudocode and implementation details of the original COMPOSE version that uses $\alpha$-shape construction to extract core supports can be seen in Algorithm \ref{alg:compose}.

COMPOSE.V1 requires the following as input: i) an SSL algorithm such as cluster and label, label propagation \cite{Zhu02}, or semi-supervised support vector machines \cite{Bennett99} with relevant free parameters; and ii) a CSE algorithm, i.e., $\alpha$ shape creation algorithm with parameters $\alpha$-shape detail level, $\alpha$, and a compaction percentage, $CP$, that represents the percentage of current labeled instances to use as core supports. The algorithm is seeded with initial labeled data $D_{init}$ in step \ref{alg:compose:step:labeled}. COMPOSE starts by receiving $N$ unlabeled instances $U^t$ in  each time-step. The SSL algorithm is then trained using the current unlabeled and labeled instances, which returns an hypothesis $h^t$ that classifies all unlabeled instances of the current time-step in step \ref{alg:compose:step:hypothesis}. The hypothesis is then used to generate a combined set of data, $D^t$, in step \ref{alg:compose:step:combine}, and the combined data for each class is used as the input for the CSE routine in step \ref{alg:compose:step:cse}. The resulting core supports $CS_c$, for each class $c$, are appended to be used as current labeled data in the next time-step in step \ref{alg:compose:step:append}

\begin{algorithm}[H]
 \caption{\textbf{COMPOSE.V1}}
 \label{alg:compose}
 \begin{algorithmic}[1]
 \renewcommand{\algorithmicrequire}{\textbf{Inputs:}}
 \REQUIRE SSL algorithm - \textbf{SSL} with relevant free parameters;
 CSE algorithm - \textbf{CSE}; $\alpha$-shape detail level-$\alpha$
 Compaction percentage - \textbf{CP}
  \STATE Receive initial labeled data $ D_{init} = \lbrace x_i;y_i \rbrace$ ; $i = 1,...,M$ ; $x\in X ; y \in Y=\lbrace1,...,c\rbrace$
  \newline Set $L^0$ = $\left\{\textnormal{$x_i^t$ }\right\}$ ; initial instances
  \newline Set $Y^0$ = $\left\{\textnormal{$y_i^t$}\right\}$ ; corresponding labels of initial instances\label{alg:compose:step:labeled}
  \FOR {$t = 0,1,....$ }
  \STATE Receive unlabeled data $U^t$ = $\left\{\textnormal{$x_u^t$ $\in$ $X$ , $u = {1,...,N}$}\right\}$
  \STATE Run \textbf{SSL} with $L^t$ , $Y^t$, and $U^t$
  \newline to obtain hypothesis, $h^t:$ X $\rightarrow$ Y \label{alg:compose:step:hypothesis}
  \STATE Let $D^t = \left\{ (x_l^t,y_l^t): x \in L^t \forall l \right\} \cup$ 
  \newline $\left\{ (x_u^t,h^t(x_u^t)): x \in U^t \forall u \right\} $ \label{alg:compose:step:combine}
  \STATE Set $L^{t+1} = \emptyset , Y^{t+1} = \emptyset$
  \FOR { each class $c = 1,2,...., C$ }
  \STATE Run \textbf{CSE} with $CP$ , $\alpha$ and $D_c^t$ 
  \newline to extract core supports, $CS_c$ \label{alg:compose:step:cse}
  \STATE Add core supports to labeled data
  \newline $L^{t+1} = L^{t+1} \cup CS_c$ 
   \newline $Y^{t+1} = Y^{t+1} \cup \left\{y_u: u \in [|CS_c|], y = c\right\}$  \label{alg:compose:step:append}
  \ENDFOR
  \ENDFOR
 \end{algorithmic}
 \end{algorithm}

\subsubsection{COMPOSE.V2 (COMPOSE With Gaussian Mixture Model (GMM) or Any Density Estimation Technique)}
One of the central processes of COMPOSE is the core support extraction, where the algorithm predicts which data instances of the current environment will be useful and relevant for classification in future time-steps, where the underlying data distributions may have changed. In the original version of COMPOSE, $\alpha$-shape construction is used for this process, but $\alpha$-shape construction is a computationally very expensive process, especially when the dimensionality of the data increases. This is because $\alpha$-shape construction requires Delaunay tessellation of the data, and the algorithm used for this purpose is the Quickhull algorithm \cite{Dyer14}. This algorithm is of order $O(n^{(d+1)/2})$ where $n$ is the number of observations and $d$ is the dimensionality of the data. Hence, the algorithm is exponential in dimensionality. In order to reduce the computational complexity of the algorithm, we make use of the fact that the goal of the CSE is to extract the labeled data from each class by creating an object or shape around the data and by compacting that object. This process is essentially equivalent to density estimation.Therefore, more efficient density estimation techniques can be used. One such approach is Gaussian Mixture Model (GMM), though any other density estimation technique can also be used here such as Parzen windows or kNN. We observe that GMM are significantly more computationally efficient than $\alpha$-shape. The Gaussian mixture model (GMM) is a probabilistic model that describes the data as a mixture of unimodal Gaussian distributions, and tries to fit $K$ Gaussians to the data $X$ where $K$ is a user specified parameter. The probability density function is the weighted sum of the $K$ Gaussians as given by the following equation,
\begin{equation}
{p(\theta) = \sum_{k=1}^K \pi_k \mathcal{N}(\mu_k , \Sigma_k)}
\end{equation}
where $\theta$ is the set of parameters describing the entire model, $\mu_k, \Sigma_k, \pi_k$ are the mean, covariance, and mixing coefﬁcient (i.e., prior probability) of each Gaussian component respectively.

The major advantage of using GMMs is that GMMs are significantly more computationally efficient than $\alpha$-shapes, particularly when $d$ is large. The computational complexity of the EM procedure for GMMs is difficult to quantify, because it is an iterative procedure, but it has been shown that the E-step and the M-step are of the order $O(NKd + NK)$ and $O(2NKd)$, respectively, for each iteration, where $N$ is the number of observations, $K$ is the number of mixture components and $d$ is the dimensionality. Our results in chapter 5 confirms that the GMM approach is indeed substantially faster than constructing $\alpha$-shapes for any given dimensionality and data cardinality.  
The pseudocode and implementation detail of COMPOSE.V2 is similar to COMPOSE.V1 with the difference of using GMM instead of the $\alpha$-shapes construction for core supports extraction module.
\subsubsection{COMPOSE.V3 (Learning Extreme Verification Latency Quickly: FAST COMPOSE)}
The third version of COMPOSE modifies the core support extraction module based on the following observation. Originally a significant overlap of class conditional distributions between consecutive time steps was thought to be the working definition of \textit{gradual / limited drift}, and hence a necessary condition for COMPOSE to work.  However, Sarnelle et al. showed in \cite{Sarnelle15} that COMPOSE can work equally well for scenarios even when there is no overlap of distributions in consecutive time steps, as long as the distance between the unlabeled data with core supports of a given class is less than the distance from the nearest core supports of any other opposing class. We refer to this condition as \textit{limited drift}, and now distinguish it from \textit{gradual drift} that does require an overlap of distributions in subsequent time-steps. As a result, we show that the condition of significant overlap (or \textit{gradual drift}) can be eliminated, and replaced with the more relaxed condition of \textit{limited drift}. We observe that for cases where there is no significant overlap, core support extraction procedure has very little impact on accuracy because it does not change centroids in any considerable amount, and clustering based SSL algorithm can easily track the drifting distributions using nearest centroids. 

Additionally, as described above, the density estimation procedure is impractical for high dimensional data due to its computational complexity. Taken together then, an obvious questions that comes to mind is whether the density estimation based core support extraction is needed at all. To answer this question, we removed the core support extraction procedure of COMPOSE entirely, and all instances labeled by the semi-supervised algorithm are then used as ``core supports," i.e., the most representative instances for the future time-steps. We call this modified version of the algorithm FAST COMPOSE \cite{7849962}. 

The pseudocode and implementation details of FAST COMPOSE are shown in Algorithm \ref{alg:fast_compose}. FAST COMPOSE only requires an SSL algorithm with its relevant free parameters as an input. The algorithm begins by receiving $M$ initially labeled instances, $L^0$, and corresponding labels $Y^0$, of $C$ classes in step \ref{alg:fast_compose:step:1}. The algorithm then receives a new set of $N$ unlabeled instances $U^t$. The SSL algorithm is then executed given the current unlabeled and labeled instances to receive the hypothesis $h^t$ of the current time-step in step \ref{alg:fast_compose:step:hypothesis}. The hypothesis is then used to label the data for the next time-step as shown in steps \ref{alg:fast_compose:step:combine} - \ref{alg:fast_compose:step:append} of Algorithm \ref{alg:fast_compose}.
\begin{algorithm}
 \caption{\textbf{FAST COMPOSE}}
  \label{alg:fast_compose}
 \begin{algorithmic}[1]
 \renewcommand{\algorithmicrequire}{\textbf{Input:}}
 \REQUIRE SSL algorithm - \textbf{SSL} with relevant free parameters
  \STATE Receive labeled data
  \newline $L^0$ = $\left\{\textnormal{$x_l^t$ $\in$ $X$}\right\}$ ,
  \newline $Y^0$ = $\left\{\textnormal{$y_l^t$ $\in$ $Y = \left\{1,\ldots,C\right\}, l = {1,\ldots,M}$}\right\}$ \label{alg:fast_compose:step:1}
  \FOR {$t = 0,1,....$ }
  \STATE Receive unlabeled data $U^t$ = $\left\{\textnormal{$x_u^t$ $\in$ $X$ , $u = {1,...,N}$}\right\}$
  \STATE Run \textbf{SSL} with $L^t$ , $Y^t$, and $U^t$
  \newline to obtain hypothesis, $h^t:$ X $\rightarrow$ Y
  \label{alg:fast_compose:step:hypothesis}
  \STATE Let $D^t = \left\{ (x_u^t,h^t(x_u^t)): x \in U^t \forall u \right\} $ \label{alg:fast_compose:step:combine}
  \STATE Set $L^{t+1} = \emptyset , Y^{t+1} = \emptyset$
  \FOR { each class $c = 1,2,...., C$ }
  \STATE  $CS_c$ = $\left\{\textnormal{$x:$ $x$ $\in$ $D_c^t$}\right\}$ , and add to labeled data for next time-step
  \newline $L^{t+1} = L^{t+1} \cup CS_c$ 
   \newline $Y^{t+1} = Y^{t+1} \cup \left\{y_u: u \in [|CS_c|], y = c\right\}$   \label{alg:fast_compose:step:append}
  \ENDFOR
  \ENDFOR
 \end{algorithmic}
 \end{algorithm}

\subsection{Stream Classification Algorithm Guided by Clustering (SCARGC)}
SCARGC is a clustering-based algorithm proposed by Souza et al \cite{Souza15_SCARGC} to deal with extreme verification latency problem, that repeatedly clusters unlabeled input data, and then classifies the clusters using the labeled clusters from the previous time-step. SCARGC also makes several assumptions: 
\begin{enumerate}
\item A small amount of labeled data is available initially to define the problem;
\item The drift is gradual / incremental, which allows tracking of the classes with only unlabeled information. Incremental drift assumption as used in SCARGC requires significant overlap between class distributions in subsequent time steps and short intervals of time;
\item The number of classes is known and fixed ahead of time. 
\end{enumerate}
Given the aforementioned assumptions, the algorithm builds an initial classification model using the available labeled data from $c$ classes, and then divide the initial labeled data into $k \geq c$ clusters where $k$ is a user-selected free parameter. If user selects $k=c$, SCARGC uses $c$ classes as initial clusters, otherwise a clustering subroutine finds clusters and associates each cluster with one class. Souza denotes this initial set of $k$ clusters as $C^0={C_1^0,C_2^0,…….,C_k^0}$. As new unlabeled data are received, the algorithm stores each example in a pool, and predicts its label using the initial classification model. After a fixed number of examples, also pre-determined by the user, are received and stored in the pool, the pool of examples is clustered into $k$ clusters in the same way as initial labeled data are clustered, i.e., by using $c$ classes as initial clusters if $k=c$, otherwise running a clustering subroutine to associate each cluster with one class. The new set of clusters are denoted as $C^1={C_1^1,C_2^1,……,C_k^1}$. Each new cluster $C_i^1 \in C^1$ is then associated with (linked to) one of the previous clusters $C_j^0 \in C^0$ to assign each cluster to one class. The classification model is updated using the recently labeled examples. The algorithm then repeats the loop, alternating between clustering and classification. The labels are decided by associating clusters $C^t$ in the current iteration with the labels of clusters $C^{t-1}$ from the previous iteration. The mapping between the clusters is performed by centroid similarity between current and previous iterations using Euclidean distance. Given the current centroids from the most recent unlabeled clusters and past centroids from the previously labeled clusters, one-nearest neighbor algorithm (or support vector machine) is used to label the centroid from current unlabeled clusters. 

SCARGC is computationally efficient, but its performance is highly dependent on the clustering phase. It also requires some prior knowledge such as the number of classes and the number of modes for each class in the data, the latter of which may limit the use of this algorithm when such information is not available. 
The pseudocode for SCARGC algorithm is given in Algorithm \ref{alg:scargc}

\begin{algorithm}[H]
 \caption{\textbf{SCARGC}}
 \label{alg:scargc}
 \begin{algorithmic}[1]
 \renewcommand{\algorithmicrequire}{\textbf{Inputs:}}
 \REQUIRE Initial training data $D_{init}$, maximum pool size $N$, number of clusters $k$;
  \STATE Receive initial labeled data $ D_{init} = \lbrace x_i;y_i \rbrace$ ; $i = 1,...,M$ ; $x\in X ; y \in Y=\lbrace1,...,c\rbrace$
 \STATE Build initial classifier $\phi$ using $D_{init}$
 \STATE Run $k$-means clustering algorithm to divide the data into $k$ clusters; $\lbrace C^t = C_1^t, C_2^t,...,C_k^t\rbrace$ and associate each cluster with one of the $c$ classes  \label{alg:scargc:step:clustering}
\STATE Start receiving new unlabeled examples from unlabeled data stream $U$ = $\left\{\textnormal{$x_u$ $\in$ $X$}\right\}$ 
 \label{alg:scargc:step:unlabel}
\STATE Store the next batch of $N$ examples in a pool
\STATE Predict labels of stored examples using classifier $\phi$ as $D_{new} = \lbrace x_u ; \phi(x_u)\rbrace ; u = 1, ... ,N$
\STATE Run $k$-means clustering algorithm on $D_{new}$ to obtain $\lbrace C^{t+1} = C_1^{t+1}, C_2^{t+1},...,C_k^{t+1}\rbrace$
\STATE Establish a mapping between current and previous clusters: the current clusters $C^{t+1}$ are associated to previous clusters $C^t$ by measuring similarity between their centroids $q_t^i; i= \lbrace 1,...,k\rbrace$ using Euclidean distance, i.e., $Dist(q_t,q_{t+1})$ where $Dist$ represents Euclidean distance 
\STATE Assign current centroid $q_{t+1}^i$ the label $\hat{y}_i$ which is same label $y_i$ of the closest past centroid $q_t^i$ 
\STATE The current dataset now has the updated correct labels from the previous step as
$D_{t+1} = \lbrace x_u ; \hat{y}_u)\rbrace ; u = 1, ... ,N$ 
\STATE Update the initial classifier $\phi$ using $D_{t+1}$
\STATE Go to step \ref{alg:scargc:step:unlabel} and repeat 
\end{algorithmic}
 \end{algorithm}
 
\subsection{Micro-Cluster for Classification (MClassification)}
 Souza et al. also proposed \textit{MClassification}, an algorithm that uses the idea of micro clusters (MC) \cite{Souza15_MClassification} to adapt to the changes in the data over time, and learn the concepts under extreme verification latency. A Microcluster (MC) is a compact representation of the data points $\vec{x_i}; i = \lbrace 1,...,N\rbrace$, that includes the sufficient statistics of the data and are represented in triplets $(N,\vec{LS},\vec{SS})$, where $N$ is the number of data points in the cluster,  $\vec{LS}$ is the linear sum of $N$ data points represented as $\vec{LS} = \lbrace \vec{x_1} + \vec{x_2} + ..... + \vec{x_n}\rbrace$, and $\vec{SS}$ is the square sum of data points represented as $\vec{SS} = \lbrace \vec{x_1}^2 + \vec{x_2}^2 + ..... + \vec{x_n}^2\rbrace$. Thus a MC summarizes the information about the set of $N$ data points, from which we can calculate the centroid and radius of the MC using the following equations
\begin{equation}
centroid = \frac{\vec{LS}}{N}
\end{equation}
\begin{equation}
Radius = \sqrt[]{\frac{\vec{SS}}{N}-(\frac{\vec{LS}}{N})^2}
\end{equation}
Although MC is efficient and appropriate for data streaming problems, the authors observe that MC representation has been commonly used in clustering problems. In order to use MC to classify evolving data streams, the authors modify the representation to store information about the class of data points, thus their representation is a 4-tuple $(N,\vec{LS},\vec{SS},y)$, where $y$ is the label for a set of data points. The working of the algorithm is presented below. 

The algorithm begins by receiving the initial labeled data $D_{init}$, using which it builds a set of labeled MCs, where each MC has information about only one example. The algorithm then starts receiving the unlabeled data stream. A label $\hat{y_t}$ is then predicted for each example $\vec{x_t}$ from the stream based on its nearest MC, computed with respect to Euclidean distance in the classification phase.
The example $\vec{x_t}$ is added to its corresponding nearest MC, say $MC_N$. Now the updated radius of $MC_N$ is computed and the algorithm checks if the updated radius of $MC_N$ exceeds the maximum micro-cluster radius threshold $r$ defined by the user. If the radius does not exceed the threshold $r$, the example $\vec{x_t}$ remains added in $MC_N$ and its updated centroid is also computed. The centroid position of the updated MC, i.e., $MC_N$ is therefore slightly moved in direction of the newly emerging concept of the class for new example added. On the other hand, if the radius exceeds the threshold, a new MC say $MC_N'$ carrying the predicted label $\hat{y_t}$ is created to allocate the new example $\vec{x_t}$. The process is repeated for each newly received unlabeled example. The pseudocode for MClassification algorithm with the implementation details is provided in Algorithm \ref{alg:mclass}.
\begin{algorithm}[H]
 \caption{\textbf{MClassification}}
 \label{alg:mclass}
 \begin{algorithmic}[1]
 \renewcommand{\algorithmicrequire}{\textbf{Inputs:}}
 \REQUIRE Maximum micro-cluster radius $r$;
  \STATE Receive initial labeled data $ D_{init} = \lbrace x_i;y_i \rbrace$ ; $i = 1,...,T$ ; $x\in X ; y \in Y=\lbrace1,...,c\rbrace$
 \STATE Build $T$ micro-clusters as $MC_i = (N_i,LS_i,SS_i,y_i) ; i = 1,...,T$ where $N$ = number of data points ; $LS = \sum_{j=1}^N x_j$ ; $SS = \sum_{j=1}^N (x_j)^2$
 \STATE Calculate sufficient statistics of each micro-cluster as follows $centroid_i = \frac{\vec{LS_i}}{N_i} ; Radius_i = \sqrt[]{\frac{\vec{SS_i}}{N_i}-(\frac{\vec{LS_i}}{N_i})^2}$ 
\STATE Receive one new unlabeled example $\vec{x}_t$ from the unlabeled data stream   
\newline $U = \lbrace x_u \in X \rbrace$ \label{alg:mclass:step:unl_ex}
\STATE Measure distance between $\vec{x_t}$ and each micro-cluster centroids $centroid_i; i= \lbrace 1,...,T\rbrace$ i.e. $Dist(centroid_i,\vec{x_t})$ to find closest micro-cluster say $MC_N$, where $Dist$ represents the Euclidean distance
\STATE Assign label of $MC_N$ i.e. $\hat{y_t}$ to classify example $\vec{x_t}$
\STATE Add example $\vec{x_t}$ to $MC_N$ and compute its sufficient statistics $radius_N$ ; and  $centroid_N$
\IF{$radius_N > r$}
\STATE Create a new micro-cluster for example $\vec{x_t}$ say $MC_N' = (N_N' , LS_N' , SS_N' , \hat{y_t})$
\ELSE
\STATE Add example $\vec{x_t}$ to $MC_N$ and update its statistics as $(\vec{LS}_N) \leftarrow (\vec{LS}_N) + \vec{x_t} ;  (\vec{SS}_N) \leftarrow (\vec{SS}_N)+ (\vec{x_t})^2 ; N_N \leftarrow N_N + 1 $ 
\ENDIF
\STATE Go to step \ref{alg:mclass:step:unl_ex} and repeat
 \end{algorithmic}
 \end{algorithm}

\subsection{LEVEL\textsubscript{IW}: \underline{L}earning \underline{E}xtreme \underline{VE}rification \underline{L}atency \\With \underline{I}mportance \underline{W}eighting}
LEVEL\textsubscript{IW} is based on the observation that importance weighting based domain adaptation used for covariate shift and concept drift problems are related, though algorithms for each make different assumptions. Concept drift problems typically assume at least a gradual (or at least limited) drift assumption, but do not require stationary posteriors or shared support while, covariate shift assumes that the class conditional distributions at consecutive time steps share support and posterior distributions do not change. 

More specifically, authors of LEVEL\textsubscript{IW} observed that COMPOSE originally assumed a significant distribution overlap at consecutive time steps, allowing instances lying in the center of the feature space to be used as the most representative labeled instances from current time step to help label the new data at the next time step. Such an assumption is also inherent in importance weighting based domain adaptation, but only for a single time step with mismatched train and test data distributions. They therefore explore importance weighting not for a single time step matching training / test distributions, but rather matching distributions between two consecutive time steps, and estimate the posterior distribution of the unlabeled data using importance weighted least squares probabilistic classifier (IWLSPC) \cite{hachiya2012importance}. The estimated labels are then iteratively used as the training data for the next time step. They call this algorithm LEVEL\textsubscript{IW}: \underline{L}earning \underline{E}xtreme \underline{VE}rification \underline{L}atency with \underline{I}mportance \underline{W}eighting. The pseudocode and implementation details of this approach are described below and summarized in Algorithm \ref{alg:LEVEL_iw}.
LEVEL\textsubscript{IW} takes advantage of the importance weighted least squares probabilistic classifier (IWLSPC) as a subroutine \cite{hachiya2012importance}, and hence serves as a wrapper approach.  

\begin{algorithm}[H]
 \caption{\textbf{LEVEL\textsubscript{IW}}}
 \label{alg:LEVEL_iw}
 \begin{algorithmic}[1]
 \renewcommand{\algorithmicrequire}{\textbf{Inputs:}}
 \REQUIRE Importance weighted least squares probabilistic classifier - \textbf{IWLSPC}; Kernel bandwidth value $\sigma$

\STATE At $t=0$, receive initial data $\textbf{x} \in X$ and the corresponding labels $ y \in Y = {1,\ldots,C}$. 
\newline Set $\textbf{x}_{te}^{t=0}$ = $\textbf{x}$
\vspace{0.1cm}
\newline Set $y_{te}^{t=0}$ = $y$
\FOR {$t = 1,...., $ }
  \STATE Receive new unlabeled test data ${\textbf{x}_{te}^t \in X}$
  \STATE Set $\textbf{x}_{tr}^t$ = $\textbf{x}_{te}^{t-1}$
  \vspace{0.1cm}
  \STATE Set $y_{tr}^t$ = $y_{te}^{t-1}$
  \STATE Call \textbf{IWLSPC} with $ \textbf{x}_{tr}^t, \textbf{x}_{te}^t, y_{tr}^t$, and $\sigma$ to estimate $y_{te}^{t}$ 
  \ENDFOR
 \end{algorithmic}
 \end{algorithm}
\vspace{0.3in}

Initially, at  $t=0$, LEVEL\textsubscript{IW} receives data $\textbf{x}$ with their corresponding labels $y$, initializes the test data $\textbf{x}_{te}^{t=0}$ to initial data $\textbf{x}$ received, and sets their corresponding labels $y_{te}^{t=0}$ equal to the initial labels $y$. Then, the algorithm iteratively processes the data, such that at each time step $t$, a new unlabeled test dataset $\textbf{x}_{te}^t$ is first received, the previously unlabeled test data from previous time step $\textbf{x}_{te}^{t-1}$, which is now labeled by the IWLSPC subroutine, becomes the labeled training data $\textbf{x}_{tr}^t$ for the current time step, and similarly the labels $y_{te}^{t-1}$ obtained by IWLSPC during the previous time step become the labels of the current training data $\textbf{x}_{tr}^t$. The training data at the current time step $\textbf{x}_{tr}^t$, the corresponding label information at the current time step $y_{tr}^t$, the kernel bandwidth value $\sigma$ and the unlabeled test data at the current time step $\textbf{x}_{te}^t$ are then passed onto the IWLSPC algorithm, which predicts the labels $y_{te}^t$ for the test unlabeled data. The entire process is then iteratively repeated.

We also note that two other algorithms are also proposed to work in the EVL setting more recently; these are called TRACE \cite{arostegi2018concept}, which tracks the trajectory of the clusters over time using some trajectory prediction algorithm for instance Kalman filter, instead of tracking clusters using unsupervised learning algorithms as done in COMPOSE and SCARGC, and Affinity-based COMPOSE \cite{razavi2019novelty}, which is based on COMPOSE with a slight modification in the core support extraction module. Affinity based COMPOSE uses only those samples from the previous timestep as the labeled information which has the highest similarity scores with the unlabeled samples at current time step, computed from the affinity matrix. These algorithms are useful contribution to the literature however, we observe that TRACE does not show any statistically significant improvement from the other algorithms already proposed in the literature and Affinity based COMPOSE only shows the incremental improvement over FAST COMPOSE in some of the experiments designed by the authors of the paper. Due to these reasons we do not include these two algorithms in our experiments.

\section{Experiments $\&$ Results}
We analyze the algorithms' behavior from three different perspectives: the average classification accuracy shown in Table \ref{table:accuracy}, computational complexity of these algorithms as measured in runtime on a fixed system shown in Table \ref{table:exec_time}, and a more detailed parameter sensitivity based analysis shown in Tables \ref{table:accuracy_scargc}, \ref{table:accuracy_mclass}, \ref{table:accuracy_compose}, and \ref{table:accuracy_level_iw}. Our analyses here include SCARGC, MClassification, COMPOSE and LEVEL\textsubscript{IW}. Arbitrary sub-population tracker (APT) was not included in the analyses, as this algorithm's steep computational complexity was prohibitive on running of some of the larger datasets. This behavior of APT was also previously reported in \cite{Dyer14}, even on a simple bi-dimensional problem. 
The analysis of this algorithm in \cite{Dyer14}, when originally compared to COMPOSE  also revealed another significant shortcoming -- that APT requires all modes of the data distribution to be present at the initialization, and hence can not accommodate scenarios where a distribution splits into multiple modes or vice versa over time. Taken together, then, these two concerns rendered APT to be less competitive compared to other algorithms in real world scenarios and hence was not included in further analysis. The reason we described the working of APT algorithm in detail in section II, is that it was the very first algorithm that introduced this problem of extreme verification latency to the computational intelligence research community, and thus motivated the other researchers in the field to propose efficient algorithms to work in the EVL setting.

The results discussed below are organized by the algorithms, discussing the observations made for each algorithm under evaluation in comparison to others. 

\begin{table*}[h]
\renewcommand{\arraystretch}{1.3}
\caption[Average classification accuracy]{\textit{Average classification accuracy}}
\label{table:accuracy}
\centering
\resizebox{\textwidth}{!}{\begin{tabular}{|c|c|c|c|c|c|c|c|}
  \hline
DATASETS & COMPOSE ($\alpha$-shape) & COMPOSE (GMM) & FAST COMPOSE & SCARGC (1-NN) & SCARGC (SVM) & MClassification & LEVEL\textsubscript{IW}  \\
\hline
1CDT & 99.96(2) & 99.85(5) & 99.97(1) & 99.69(7)  & 99.72(6) & 99.89(4) &  99.92(3)\\   
\hline                                                                              
1CHT & 99.60(2) & 99.34(6) & 99.57(3) & 99.69(1) & 99.27(7) & 99.38(5) & 99.52(4) \\       
\hline                                                                              
1CSurr & 90.95(5) & 89.72(6) & 95.64(1) & 94.53(3) & 94.99(2) & 85.15(7) & 91.30(4) \\       
\hline                                                                              
2CDT & 96.58(1) & 95.92(2) & 95.17(4) & 87.71(6) & 87.82(5) & 95.23(3) & 58.32(7) \\        
\hline 
2CHT & 90.39(1) & 89.63(2) & 89.41(3) & 83.62(5) & 83.39(6) & 87.93(4) & 52.15(7) \\
\hline                  
4CE1CF & 93.92(5) & 93.90(6) & 93.95(4) & 94.04(3) & 92.79(7) & 94.38(2) & 97.74(1) \\     
\hline                                                                              
4CR & 99.99(2.5) & 99.99(2.5) & 99.99(2.5) & 99.96(6) & 98.94(7) & 99.98(5) & 99.99(2.5) \\         
\hline                                                                              
4CRE-V2 & 92.59(1) & 92.30(3) & 92.46(2) & 91.34(6) & 91.46(5) & 91.59(4) & 24.10(7) \\    
\hline                                                                              
FG\_2C\_2D & 87.90(6) & 95.50(5) & 95.58(3) & 95.51(4) & 95.60(2) & 62.48(7) & 95.71(1) \\           
\hline                                                                              
GEARS\_2C\_2D & 90.98(7) & 95.83(3) & 91.26(6)  & 95.99(2) & 95.81(4) &  94.73(5) & 97.74(1)\\       
\hline                                                                              
MG\_2C\_2D & 93.12(2) & 93.20(1) & 93.02(3) & 92.92(5) & 92.94(4) & 80.58(7) & 85.44(6) \\         
\hline                                                                              
UG\_2C\_2D & 95.63(3) & 95.71(1) & 95.61(5) & 95.65(2) & 95.62(4) & 95.28(6) & 74.34(7) \\   
\hline                                                                              
UG\_2C\_3D & 94.92(3) & 95.20(1) & 95.12(2) & 94.83(5) & 94.91(4) & 94.72(6) & 64.69(7) \\            
\hline                                                                              
UG\_2C\_5D & 92.07(2) & 92.13(1) & 91.99(3)  & 91.38(4) & 90.94(6) & 91.25(5) & 80.17(7) \\  
\hline                                                                              
keystroke & 84.31(7) & 87.21(5) & 85.92(6) & 88.07(3.5) & 88.07(3.5) & 90.62(1) & 90.56(2) \\      
                           \hline
                           \hline
Average Rank (lower is better)  & 3.2813 & 3.4688 & 3.1563 & 4.1563 & 4.8438 & 4.5000 & 4.5938 \\ 
							\hline
\end{tabular}}
\end{table*}

\begin{table*}[t]
\renewcommand{\arraystretch}{1.3}
\caption[Average execution time (in seconds)]{\textit{Average execution time (in seconds)}}
\label{table:exec_time}
\centering
\resizebox{\textwidth}{!}{\begin{tabular}{|c|c|c|c|c|c|c|c|}
  \hline
DATASETS & COMPOSE ($\alpha$-shape) & COMPOSE (GMM) & FAST COMPOSE & SCARGC (1-NN) & SCARGC (SVM) & MClassification & LEVEL\textsubscript{IW}  \\
\hline
1CDT & 19.18(6) & 4.21(3) & 1.15(1) & 10.20(4)  & 2.50(2) & 64.75(7) &  15.02(5)\\   
\hline                                                                              
1CHT & 19.76(6) & 4.04(3) & 1.17(1) & 10.76(4) & 3.29(2) & 62.36(7) & 15.34(5) \\       
\hline                                                                              
1CSurr & 72.84(6) & 7.32(2) & 2.53(1) & 51.78(5) & 16.40(3) & 220.49(7) & 43.83(4) \\       
\hline                                                                              
2CDT & 20.21(6) & 2.89(2) & 1.46(1) & 10.00(4) & 3.34(3) & 62.48(7) & 15.71(5) \\        
\hline 
2CHT & 19.59(6) & 3.55(3) & 1.41(1) & 10.09(4) & 2.89(2) & 60.77(7) & 15.79(5) \\
\hline                  
4CE1CF & 241.16(6) & 44.14(2) & 8.41(1) & 210.97(5) & 134.56(3) & 775.59(7) & 137.82(4) \\     
\hline                                                                              
4CR & 213.51(6) & 55.90(2) & 12.04(1) & 91.22(4) & 56.22(3) & 608.00(7) & 148.32(5) \\         
\hline                                                                              
4CRE-V2 & 216.55(5) & 34.82(2) & 6.44(1) & 280.27(6) & 41.51(3) & 641.46(7) & 147.81(4) \\    
\hline                                                                              
FG\_2C\_2D & 229.34(5) & 16.04(2) & 3.80(1) & 587.19(6) & 54.58(3) & 870.12(7) & 185.77(4) \\           
\hline                                                                              
GEARS\_2C\_2D & 237.24(5) & 14.45(2) & 2.50(1)  & 609.95(7) & 26.91(3) &  497.87(6) & 186.42(4)\\       
\hline                                                                              
MG\_2C\_2D & 228.96(5) & 15.38(2) & 4.26(1) & 583.76(6) & 53.44(3) & 740.75(7) & 190.81(4) \\         
\hline                                                                              
UG\_2C\_2D & 115.30(5) & 16.92(2) & 3.45(1) & 152.24(6) & 23.27(3) & 362.48(7) & 72.69(4) \\   
\hline                                                                              
UG\_2C\_3D & 936.18(7) & 15.64(2) & 2.60(1) & 747.96(5) & 62.28(3) & 881.07(6) & 176.53(4) \\            
\hline                                                                              
UG\_2C\_5D & 2138.39(7) & 15.97(2) & 2.65(1)  & 849.03(5) & 265.92(4) & 977.53(6) & 176.84(3) \\  
\hline                                                                              
keystroke & 31761.70(7) & 2.02(4) & 1.16(3) & 0.82(2) & 0.68(1) & 6.62(6) & 2.30(5) \\      
                           \hline
                           \hline
Average Rank (lower is better)  & 5.8125 & 2.3750 & 1.1250 & 4.9375 & 2.6875 & 6.7500 & 4.3125 \\ 
							\hline
\end{tabular}}
\end{table*} 

\subsection{Analysis of Three Versions of COMPOSE}
\subsubsection{Accuracy Comparison}
Average accuracy results comparing all three versions of COMPOSE (COMPOSE with $\alpha$-shapes, with GMM and FAST COMPOSE), do not show any significant difference among them, or among any of the other algorithms as shown in Table \ref{table:significance_accuracy}. We do observe, however, that FAST COMPOSE -- while not quite with statistical significance at 0.05 level -- does perform consistently better on most datasets compared to all other algorithms, and provides the lowest overall average rank (lower rank is better in performance, rank 1 is the best algorithm and rank 7 is the worst algorithm). 
\subsubsection{Computational Complexity Comparison}
Computational complexity (as measured in seconds for runtime) among the three versions of COMPOSE as well as other algorithms also provide some useful and interesting results. As shown in Table \ref{table:significance_exec_time}, COMPOSE ($\alpha$-shape) is found to be the second worst algorithm in terms of computational complexity after MClassification, and performs significantly worse than all other algorithms except SCARGC (1-NN), MClassification and LEVEL\textsubscript{IW} (with no significant difference among the last four). For COMPOSE ($\alpha$-shape), the curse of dimensionality is the biggest bottleneck as can be seen from the significantly large computation time it takes for two datasets with even modestly high dimensionality: a 5-dimensional dataset $UG\_2C\_5D$ and the 10-dimensional real world dataset $keystroke$. We can easily see that the computational complexity of COMPOSE ($\alpha$-shape) increases exponentially with dimensionality, and therefore is impractical to use for large dimensional datasets. With respect to execution time, COMPOSE (GMM) shows significant improvement over COMPOSE ($\alpha$-shape), SCARGC (1-NN), and MClassification as seen in Table \ref{table:significance_exec_time}. FAST COMPOSE shows significant improvement over all other algorithms except COMPOSE (GMM) and SCARGC (SVM). We observe that FAST COMPOSE also performs consistently better on most datasets with respect to computation time, providing the lowest rank as shown in Table \ref{table:exec_time}. FAST COMPOSE thus comes out to be the fastest running algorithm to handle extreme verification latency. 
\subsubsection{Parameter Sensitivity Comparison}
In addition to classification accuracy and runtime based computational complexity, we also investigated the parameter sensitivity of these algorithms. Parameter sensitivity analysis measures the robustness of a given algorithm's performance in response to changes in the algorithm's most influential free parameters. In general, we prefer \textit{stable algorithms}, whose performances do not change wildly for modest changes in their free parameters. 

COMPOSE.V1 and COMPOSE.V2 employ two modules, namely core support extraction and semi-supervised learning (SSL), each requiring their own free-parameters. The primary free parameters for COMPOSE-V1 are $\alpha$-shape detail level $\alpha$, $\alpha$-shape compaction percentage $CP$, and the number of clusters $k$ for cluster and label SSL algorithm. COMPOSE.V2 requires the number of Gaussian mixtures components $K$, compaction percentage parameter $CP$, and number of clusters parameter $k$ for cluster and label SSL algorithm. All these parameters normally require fine tuning in order to give good results. COMPOSE.V3, i.e. FAST COMPOSE, is introduced primarily to reduce the computation complexity of the algorithm, but it also reduces the number of free-parameters by removing the core support extraction module, and hence requires only the number of clusters parameter $k$. Therefore we perform the sensitivity analysis of COMPOSE with respect to this parameter common to all three versions of COMPOSE. Table \ref{table:accuracy_compose} shows the results obtained by COMPOSE using cluster-and-label, where for each dataset, we provide the COMPOSE performance with the optimal $k$ value, as well as $k$ incorrectly chosen by just "1."  This $\pm 1$ represents the smallest possible change in $k$ around its optimal value. For example, if the optimal value is $k=4$, the three values of $k$ used for comparison are $k=3, k=4$, and $k=5.$ When optimal $k$ is two, the selection of $k=1$ is, of course, meaningless, as $k=1$ would result in all instances being classified into the same class. Hence, such cases are indicated as N/A in Table \ref{table:accuracy_compose}. We observe that the cluster-and-label is able to identify the structure in the data from few labeled instances, and it does so reasonably well even when there is overlap among the clusters. However, this performance is subject to correct choice of the number of clusters $k$ in the data, to which it tends to be rather sensitive, and in most datasets changing the value of $k$ from the optimal value even just by 1, significantly and catastrophically reduces the average accuracy for that dataset.

In summary, then, there is no statistically significance difference among any of the algorithms with respect to classification accuracy (though FAST COMPOSE consistently perform better). FAST COMPOSE and COMPOSE with GMM are significantly better in terms of runtime, and LEVEL\textsubscript{IW} appears to be more robust with respect to parameter variations among other algorithms.

\begin{table*}[t]
\renewcommand{\arraystretch}{0.9}
\caption[Statistical significance at $\alpha =  0.05$ for classification accuracy]{\textit{Statistical significance at $\alpha =  0.05$ for classification accuracy}}
\label{table:significance_accuracy}
\centering
\resizebox{\textwidth}{!}{\begin{tabular}{|c|c|c|c|c|c|c|c|}
 \hline & COMPOSE($\alpha$-shape) & COMPOSE(GMM) & FAST COMPOSE & SCARGC(1-NN) & SCARGC(SVM) & MClassification & LEVEL\textsubscript{IW}\\
 \hline COMPOSE($\alpha$-shape) & n/a & & & & & & \\
 \hline COMPOSE(GMM) & & n/a & & & & & \\
 \hline FAST COMPOSE & & & n/a & & & & \\
 \hline SCARGC(1-NN)  & & & & n/a & & & \\
 \hline SCARGC(SVM) & & & & & n/a & &\\
\hline MClassification & &  &  & &  & n/a &\\
\hline LEVEL\textsubscript{IW}  & &  & & & & & n/a
\\
\hline
\end{tabular}}
\end{table*}

\begin{table*}[t]
\renewcommand{\arraystretch}{0.9}
\caption[Statistical significance at $\alpha =  0.05$ for execution time]{\textit{Statistical significance at $\alpha =  0.05$ for execution time}}
\label{table:significance_exec_time}
\centering
\resizebox{\textwidth}{!}{\begin{tabular}{|c|c|c|c|c|c|c|c|}
 \hline & COMPOSE($\alpha$-shape) & COMPOSE(GMM) & FAST COMPOSE & SCARGC(1-NN) & SCARGC(SVM) & MClassification & LEVEL\textsubscript{IW}\\
 \hline COMPOSE($\alpha$-shape) & n/a & $\uparrow$ & $\uparrow$ &  & $\uparrow$ & & \\
 \hline COMPOSE(GMM) & $\leftarrow$ & n/a & & $\leftarrow$& & $\leftarrow$ &  \\
 \hline FAST COMPOSE & $\leftarrow$ & & n/a & $\leftarrow$ & & $\leftarrow$ & $\leftarrow$ \\
 \hline SCARGC(1-NN)  &  & $\uparrow$ & $\uparrow$ & n/a & &  & \\
 \hline SCARGC(SVM) & $\leftarrow$ & & & & n/a & $\leftarrow$ & \\
\hline MClassification & & $\uparrow$ & $\uparrow$  &  & $\uparrow$  & n/a &  $\uparrow$\\
\hline LEVEL\textsubscript{IW}  & & & $\uparrow$ & &  &$\leftarrow$ & n/a
\\
\hline
\end{tabular}}
\end{table*}

\subsection{Analysis of SCARGC}
\subsubsection{Accuracy Comparison}
We included two versions of SCARGC, one using nearest neighborhood (1NN) and the other using support vector machines (SVM), neither of which provided any significant difference over any of the other algorithms in terms of classification accuracy, as shown in table \ref{table:significance_accuracy}. 
\subsubsection{Computational Complexity Comparison}
With respect to the execution time, SCARGC (1-NN) does not show significant improvement over any algorithm, while SCARGC (SVM) shows significant improvement over COMPOSE ($\alpha$-shape), and MClassification as shown in table \ref{table:significance_exec_time}. FAST COMPOSE showed a significant improvement over SCARGC (1-NN) as previously discussed, and as can also be seen in table \ref{table:significance_exec_time}: the computational performance of FAST COMPOSE is not significantly better than SCARGC (SVM), however, FAST COMPOSE does take less computation time on almost every dataset as compared to SCARGC (SVM). 
\subsubsection{Parameter Sensitivity Comparison}
SCARGC has three input parameters, initial labeled data, pool size and the number of clusters. The authors in the paper \cite{Souza15_SCARGC} show that SCARGC is robust to the change in the values of the initial labeled data and the pool size (the number of instances in each batch evaluated by the algorithm at any given time). Therefore, we fixed and set the pool size equal to the batch size (drift interval shown in Table 1) used in all versions of COMPOSE and LEVEL\textsubscript{IW} to ensure the fairness in comparison. As with all algorithms, we also assume that the entire initial batch of the data is labeled, followed by all unlabaled data. This allows all algorithm to see the exact same data in each batch. The third parameter, the number of clusters $k$, is the more useful one to test with respect to the parameter sensitivity. For the sensitivity analysis, we followed a similar procedure as we did for COMPOSE, and we evaluated SCARGC using the optimal $k$ value, as well as $k$ incorrectly chosen by just "1", as shown in Table \ref{table:accuracy_scargc}. We observed that performance shown by SCARGC is also quite sensitive to correct choice of this parameter, as the performance drops dramatically and significantly for incorrect choices of $k$, particularly for the cases with class overlap. Overestimating the value of $k$ from its optimal value does not hurt the performance much for those datasets that do not have class overlap, though - perhaps not surprisingly - underestimating this value does negatively impact the classification accuracy.
\subsection{Analysis of MClassification}
\subsubsection{Accuracy Comparison} MClassification behaves similarly to other algorithms in terms of the classification accuracy when averaged across all datasets, and does not provide any significant difference as shown in Table \ref{table:significance_accuracy}. However, it performs worse than any other algorithms on two specific datasets i.e. $FG\_2C\_2D$, and $MG\_2C\_2D$. By looking in more depth the evolution of the classes at different time-steps, we notice that for both of these datasets there is a sudden change in the modes or clusters representing two classes of the data at some time-step, co-occurring with the overlap of classes that causes the drop in the performance. The other algorithms do see a small drop in their performance when the overlap occurs (as expected), but they do not lose track of the clusters with the sudden change in the positions and parameters of the distributions representing classes.
\subsubsection{Computational Complexity Comparison} With respect to the execution time, this algorithm appears to be the worst algorithm (other than APT), providing the highest rank (highest being the worst and lowest being the best) as shown in Table \ref{table:exec_time}. As shown in Table \ref{table:significance_exec_time}, MClassification takes significantly longer to run than all other algorithms, except COMPOSE with $\alpha$-shape (and perhaps APT) with which the difference is not significant.
\subsubsection{Parameter Sensitivity Comparison} From the parameter sensitivity perspective, we first note that MClassification is introduced by the same authors of SCARGC as an alternative that is claimed to use a parameter that is less sensitive and requires no prior knowledge to tune. The only parameter this algorithms uses is the maximum micro-cluster radius threshold $r$, a user-defined parameter that the authors claim is quite robust. The authors further argue that the value $r$ = 0.1 works generally well in all cases. In order to test this claim, we evaluated this algorithm on 8 different values of the parameter $r$, i.e., 0.01, 0.05, 0.1, 0.2, 0.5, 1, 1.5 and 2, whose results are given in the Table \ref{table:accuracy_mclass}. For each of the datasets, three different values of $r$ were used, representing the smallest possible value of 0.01 and largest value of $r$ among all values on which the algorithm starts seeing a drop in its performance, and the claimed default value of $r$ = 0.1. We observed that for all datasets except $MG\_2C\_2D$, the lower values of 0.01 and 0.05 do not make any difference to the performance from the optimal value. However, the performance does not remain consistent when the values greater than the optimal value are used: increasing the threshold value decreases the performance. The performance decreases more dramatically for the datasets that possess significant class overlap. 

\section{Analysis of LEVEL\textsubscript{IW}}
\subsubsection{Accuracy Comparison} The average classification accuracy shown by LEVEL\textsubscript{IW} for all datasets was, as previously mentioned, not statistically significantly different from the remaining algorithms as shown in Table \ref{table:significance_accuracy}. However, we observe that LEVEL\textsubscript{IW} performs specifically rather poorly for datasets with significant between-class overlap, as can be seen from Table \ref{table:accuracy}. The reason for this relatively poor performance can be traced to the assumptions made by domain adaptation algorithms: the significant between-class overlap coupled with a drifting environment ultimately leads to a significant change in the posterior probability distribution $p(y|x)$ of classes, violating one of the covariate shift assumptions behind domain adaptation algorithms in general, and LEVEL\textsubscript{IW} in particular. We note that the ability of other algorithms to perform well even under significant between-class overlap is in fact due to a crucial piece of information provided to them, through one of their free-parameters. \subsubsection{Computational Complexity Comparison} With respect to the execution time, LEVEL\textsubscript{IW} is significantly slower compared to FAST COMPOSE only, as shown in Table \ref{table:significance_exec_time}.
\subsubsection{Parameter Sensitivity Comparison} Since LEVEL\textsubscript{IW} is a wrapper around an algorithm IWLSPC, the free parameters for LEVEL\textsubscript{IW} are the same as the free parameters of IWLSPC algorithm i.e. the regularization parameter $\lambda$, and the kernel bandwidth parameter $\sigma$. However, the more influential free parameter for LEVEL\textsubscript{IW} is the value of the kernel width $\sigma$ as used in Gaussian kernel, therefore we performed the sensitivity analysis for this parameter. The kernel width does not provide any direct information on the number of clusters, but rather on the overall smoothness of the decision boundaries. Such information, while not terribly useful after a complete overlap, provides more protection and less sensitivity to minor or even moderate changes in its value. To see this effect, a parameter sweep range was chosen to cover a range commonly known to work well in other algorithms that use Gaussian kernels, and include the values of  0.01, 0.1, 0.2, 0.5, 1, 1.2, 1.5, 2, and 5. In Table \ref{table:accuracy_level_iw}, we show the performance of LEVEL\textsubscript{IW} for each of the datasets with three different values of $\sigma$, representing the smallest and largest values of $\sigma$ on which the algorithm performs well, as well as an additional value in the middle of the two. We observe that LEVEL\textsubscript{IW} is surprisingly robust to such wide fluctuations of $\sigma$ values of typically five fold, and sometimes as wide as an order of magnitude difference. This outcome shows the consistent and stable performance of LEVEL\textsubscript{IW}, its most prominent advantage over remaining algorithms.

\section{Analysis on two Additional Real World Datasets}
The \textit{Keystroke} dataset that was included in all aforementioned experiments is the only real world dataset in the original benchmark. That benchmark was used in part because it was used by other algorithms, allowing a fair comparison of our results to those reported in their respective publications \cite{Krempl11,Dyer14,Souza15_SCARGC,Souza15_MClassification,7849962}. We had access to two additional datasets which we used separately, on which we evaluated all four main groups of algorithms. In this section we discuss the behavior of these algorithms on these two additional real world datasets, namely \textit{Weather} and \textit{Traffic} datasets. For this analysis, among three versions of COMPOSE, we use COMPOSE.V3 (FAST COMPOSE), because of its fewer parameter requirements and reduced computational complexity, and in general we know that it works as well or better than the previous two versions. 

\begin{figure}
\centering
\includegraphics[width=0.45\textwidth]{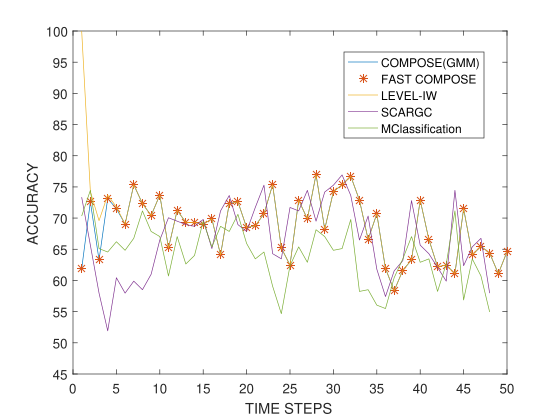}
\caption{Accuracy of algorithms on real world \textit{weather} data}
\label{fig:weather}
\end{figure}

The \textit{Weather} dataset is created by \cite{Dyer14}, and is based on the raw data obtained from the National Oceanic and Atmospheric Administration (NOAA) department. The raw data was collected over a 50-year span from Offutt Air Force Base in Bellevue, Nebraska. Eight features (temperature, dew point, sea-level pressure, visibility, average wind speed, max sustained wind speed, and minimum and maximum temperature) are used to determine whether each day experienced precipitation (rain) or not. The data set contains 18,159 daily readings of which 5,698 are \textit{rain} and the remaining 12,461 are \textit{no rain}. Hence this data has moderate class imbalance with 68.62\% of the instances belonging to class 1 while 31.38\% of the instances belonging to the other class. Data were grouped into 49 batches of one-year intervals, each containing 365 instances (days); the remaining data were placed into the fiftieth batch as a partial year. The imbalance inherent in the overall data, combined with consistent significant class overlap caused all algorithms to classify all data to one class, giving (a false sense of) accuracy of 69\% on this dataset as shown in Figure \ref{fig:weather}. Therefore, the results on this dataset are inconclusive.

\begin{figure}[h]
\centering
\vspace{0.3in}
\includegraphics[width=0.45\textwidth]{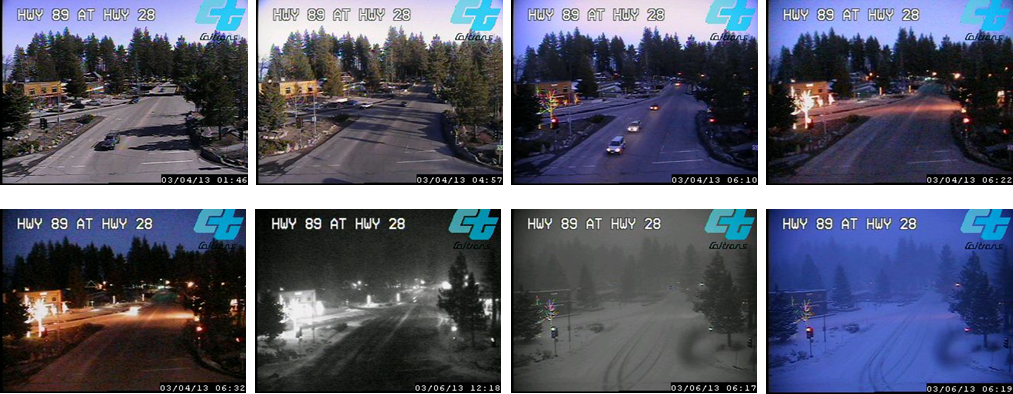}
\caption{Sample images of traffic scenes streaming from a traffic camera}
\label{fig:traffic_sample}
\vspace{0.3in}
\end{figure}

The second real dataset we use in our analysis is the \textit{Traffic} dataset, which was first introduced in \cite{hoffman2014continuous}. This dataset consists of 5,412 instances, 512 real attributes and 2 classes, representing whether a traffic intersection is busy (has cars in the intersection) or empty. The images in this dataset are captured from a fixed traffic camera continuously observing an intersection over a two-week period. Some sample images of this dataset are shown in Figure \ref{fig:traffic_sample}.

The concept drift in this dataset is due to the ambient changes in the scene that occur because of the variations in illumination, shadows, fog, snow, or even light saturation from oncoming cars, etc. We observe that this dataset also possesses imbalance but not as significant as the \textit{Weather} dataset: out of 5,412 instances, 3,168 instances (58.54\%) belong to class 1, while 2,244 instances (41.46\%) belong to the other class. While the overall data does not have significant imbalance inherent in it, dividing the data into batches does add significant imbalance to certain batches of data. The imbalance becomes increasingly more significant with the number of batches. 

\begin{figure}
\centering
\begin{subfigure}{0.5\textwidth}
\includegraphics[width=3.7 in]{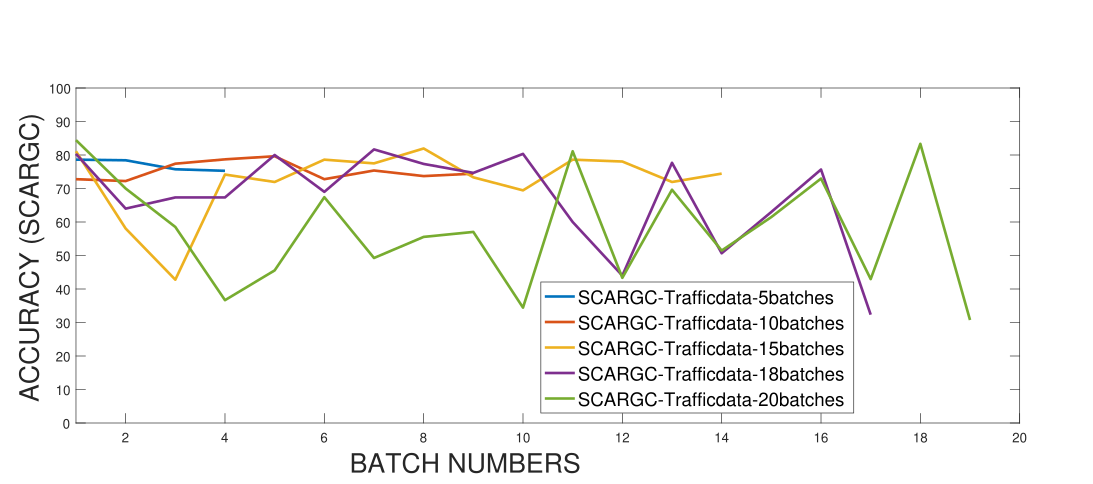}%
\hspace{-3in}
\centering\subcaption{}
\end{subfigure} 

\begin{subfigure}{0.5\textwidth} 
\includegraphics[width=3.73in]{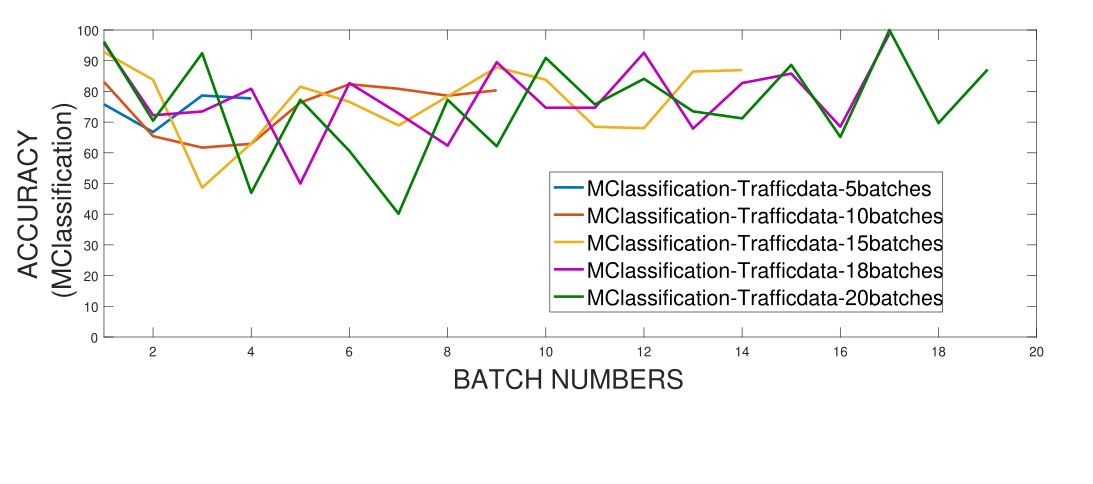}
\vspace{-0.005in}
\centering\subcaption{}
 \end{subfigure}
 
 \begin{subfigure}{0.5\textwidth}  
\includegraphics[width=3.63 in]{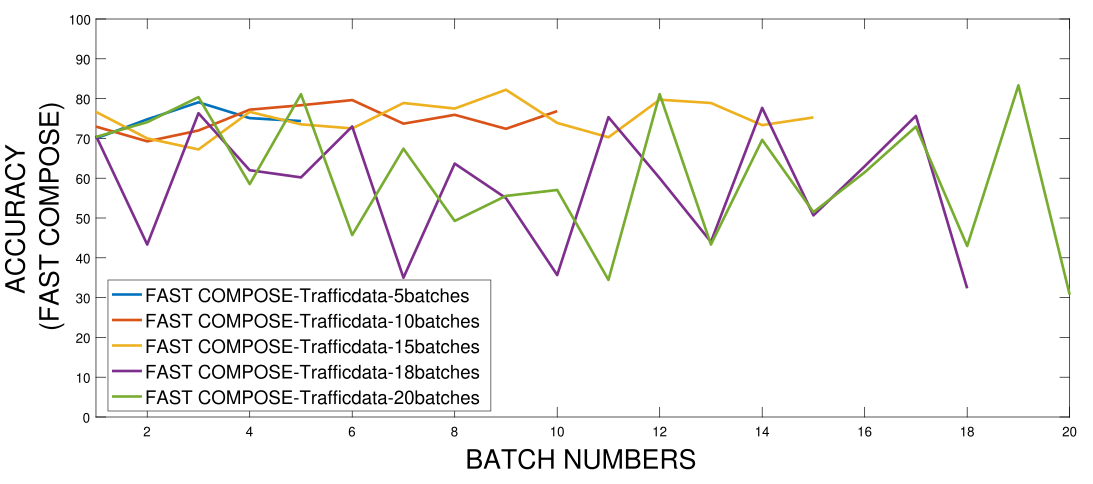}
\hspace{-3in}
\centering\subcaption{}
 \end{subfigure} 

\begin{subfigure}{0.5\textwidth} 
\includegraphics[width=3.7in]{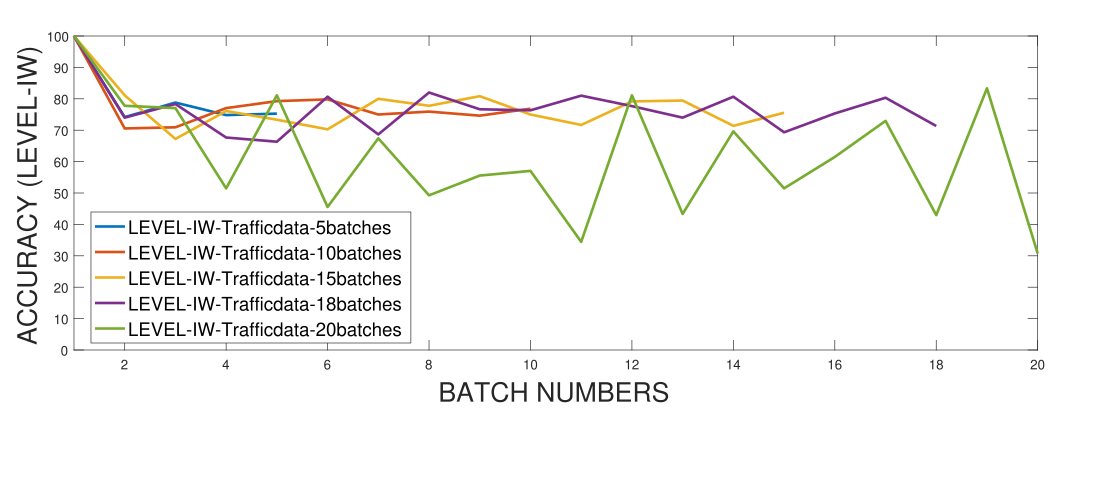}\vspace{-0.3in}
\hspace{-3in}
\centering\subcaption{}
 \end{subfigure}
 
\hspace{3in} \caption{Accuracy of algorithms on \textit{Traffic} dataset using various batch sizes (a) SCARGC performance (b) MClassification performance (c) FAST COMPOSE performance (d) LEVEL\textsubscript{IW} performance}\label{fig:traffic}
\end{figure}

Figure \ref{fig:traffic} shows the performance of each algorithm on this dataset with different number of batches, where Figure \ref{fig:traffic}(a) represents the classification accuracy of SCARGC for 5, 10, 15, 18, and 20 batches. Figure \ref{fig:traffic}(b), Figure \ref{fig:traffic}(c), and Figure \ref{fig:traffic}(d) show the same information for MClassification, FAST COMPOSE and LEVEL\textsubscript{IW}, respectively. We observe that all  algorithms show around 76\% classification accuracy, so long as the number of batches is less than or equal to 18.

The only minor exception is MClassification algorithm, which can perform equally well even if the data is divided into more than 18 batches (for instance 20 batches as seen in Figure \ref{fig:traffic}(b)). We attribute this behavior to the online nature of this algorithm, as it can process data one example or instance at a time, and hence the algorithm is not bothered by the batch size. With all other algorithms, the problem with batch size can be linked to the class imbalance:  If the data is split into 20 batches, ten  batches contain on average 68\% and 32\% imbalance among classes, while the other ten batches contain imbalance on average equal to the imbalance of the overall data i.e. 58.54\% and 41.46\%. These results further confirm a mutual shortcoming of concept drift algorithms that are asked to work under extreme verification latency that they are sensitive to class imbalance.

\begin{table*}[t]
\renewcommand{\arraystretch}{0.95}
\caption[Accuracy with three different values of k  (SCARGC)]{\textit{Accuracy with three different values of k  (SCARGC)}}
\label{table:accuracy_scargc}
\centering
\begin{tabular}{|c|c|c|c|}
  \hline
DATASETS & Reduced k (Accuracy) & Optimal k (Accuracy) & Increased k (Accuracy)\\
\hline
1CDT &  N/A & k=2 (99.72) & k=3 (99.72) \\   
\hline                                                                              
1CHT &  N/A & k=2 (99.27) & k=3 (99.22) \\       
\hline                                                                              
1CSurr &  k=4 (91.68) & k=5 (94.99) & k=6 (91.66) \\       
\hline                                                                              
2CDT &  N/A & k=2 (87.82) & k=3 (51.99) \\        
\hline 
2CHT &  N/A & k=2 (83.39) & k=3 (67.48) \\

\hline                                                                              
4CE1CF &  k=4 (2.15) & k=5 (92.79) & k=6 (49.67) \\     
\hline                                                                              
4CR &  k=3 (25.33) & k=4 (98.94) & k=5 (98.94) \\         
\hline                                                                              
                                                                        
4CRE-V2 &  k=3 (24.82) & k=4 (91.46) & k=5 (39.72)  \\    
\hline                                                     
FG\_2C\_2D &  k=3 (68.49) & k=4 (95.60) & k=5 (94.91) \\           
\hline                                                                              
GEARS\_2C\_2D &      N/A & k=2 (95.81) & k=3 (88.06)  \\       
\hline                                                                              
MG\_2C\_2D &  k=3 (64.87) & k=4 (92.94) & k=5 (82.76) \\         
\hline                                                                              
UG\_2C\_2D &  N/A & k=2 (95.62) & k=3 (57.19)  \\   
\hline                                                                              
UG\_2C\_3D &  N/A & k=2 (94.91) & k=3 (80.20)  \\            
\hline                                                                              
UG\_2C\_5D &  N/A & k=2 (90.94) & k=3 (75.08)  \\      
\hline
keystroke &  k=9 (57.43) & k=10 (88.07) & k=11 (58.07)  \\      
\hline
\end{tabular}
\vspace{0.3in}
\end{table*}

\begin{table*}[t]
\renewcommand{\arraystretch}{0.95}
\caption[Accuracy with three different values of $r$ for (MClassification)]{\textit{Accuracy with three different values of $r$ (MClassification)}
\label{table:accuracy_mclass}}
\centering
\begin{tabular}
{|c|c|c|c|}
  \hline
DATASETS & lowest $r$ (Accuracy) & Middle  $r$ (Accuracy) & Highest $r$ (Accuracy)\\
\hline
1CDT &  $r$=0.01(99.85) & $r$=0.1(99.89) & $r$=2(97.85) \\   
\hline                                                                              
1CHT &   $r$=0.01(99.23) & $r$=0.1(99.38) & $r$=2(92.97) \\       
\hline                                                                              
1CSurr &  $r$=0.01(84.80) & $r$=0.1(85.15) & $r$=0.5(48.67) \\       
\hline                                                                              
2CDT &  $r$=0.01(94.76) & $r$=0.1(95.23) & $r$=0.5(55.84) \\        
\hline 
2CHT &  $r$=0.01(86.50) & $r$=0.1(87.93) & $r$=0.5(56.37) \\

\hline                                                                              
4CE1CF &   $r$=0.01(94.59) & $r$=0.1(94.38) & $r$=2(96.21) \\     
\hline                                                                              
4CR &   $r$=0.01(99.98) & $r$=0.1(99.98) & $r$=1(23.02) \\         
\hline                                                                              
                                                                        
4CRE-V2 &   $r$=0.01(91.21) & $r$=0.1(91.59) & $r$=0.5(27.80)  \\    
\hline                                                     
FG\_2C\_2D &   $r$=0.01(59.20) & $r$=0.1(62.48) & $r$=0.2(55.84)\\           
\hline                                                                              
GEARS\_2C\_2D &  $r$=0.01(95.23) & $r$=0.1(94.73) & $r$=0.3(93.90) \\       
\hline                                                                              
MG\_2C\_2D &   $r$=0.01(51.10) & $r$=0.1(80.58) & $r$=0.2(74.41) \\         
\hline                                                                              
UG\_2C\_2D &  $r$=0.01(95.12) & $r$=0.1(95.28) & $r$=0.5(51.87) \\   
\hline                                                                              
UG\_2C\_3D &  $r$=0.01(94.57) & $r$=0.1(94.72) & $r$=0.5(52.44)  \\            
\hline                                                                              
UG\_2C\_5D &  $r$=0.01(91.31) & $r$=0.1(91.25) & $r$=1(68.17) \\      
\hline
keystroke & $r$=0.01(90.62) & $r$=0.1(76.90) & $r$=0.2(73.86)  \\      
\hline
\end{tabular}
\vspace{0.3in}
\end{table*}

\begin{table*}[t]
\renewcommand{\arraystretch}{0.95}
\caption[Accuracy with three different values of k (COMPOSE)]{\textit{Accuracy with three different values of k (COMPOSE)}}
\label{table:accuracy_compose}
\centering
\begin{tabular}{|c|c|c|c|}
  \hline
DATASETS & Reduced k (Accuracy) & Optimal k (Accuracy) & Increased k (Accuracy)\\
\hline
1CDT &  N/A & k=2 (99.85) & k=3 (99.76) \\   
\hline                                                                              
1CHT &  N/A & k=2 (99.34) & k=3 (98.72) \\       
\hline                                                                              
1CSurr &  k=3 (85.58) & k=4 (94.55) & k=5 (91.52) \\       
\hline                                                                              
2CDT &  N/A & k=2 (95.91) & k=3 (52.91) \\        
\hline 
2CHT &  N/A & k=2 (89.63) & k=3 (77.33) \\

\hline                                                                              
4CE1CF &  k=4 (78.96) & k=5 (93.90) & k=6 (94.66) \\     
\hline                                                                              
4CR &  k=3 (74.88) & k=4 (99.98) & k=5 (99.98) \\         
\hline                                                                              
                                                                        
4CRE-V2 &  k=3 (25.13) & k=4 (92.30) & k=5 (22.78)  \\    
\hline                                                     
FG\_2C\_2D &  k=3 (68.91) & k=4 (95.50) & k=5 (95.44) \\           
\hline                                                                              
GEARS\_2C\_2D &      N/A & k=2 (95.82) & k=3 (87.99)  \\       
\hline                                                                              
MG\_2C\_2D &  k=3 (65.32) & k=4 (93.20) & k=5 (92.07) \\         
\hline                                                                              
UG\_2C\_2D &  N/A & k=2 (95.71) & k=3 (56.28)  \\   
\hline                                                                              
UG\_2C\_3D &  N/A & k=2 (95.20) & k=3 (91.46)  \\            
\hline                                                                              
UG\_2C\_5D &  N/A & k=2 (92.12) & k=3 (88.03)  \\      
\hline
keystroke &  k=9 (68.62) & k=10 (87.21) & k=11 (81.56)  \\      
\hline
\end{tabular}
\vspace{0.3in}
\end{table*}

\begin{table*}[t]
\renewcommand{\arraystretch}{0.95}
\caption[Accuracy with three different values of sigma (LEVEL\textsubscript{IW})]{\textit{Accuracy with three different values of sigma (LEVEL\textsubscript{IW})}}
\label{table:accuracy_level_iw}
\centering
\begin{tabular}{|c|c|c|c|}
  \hline
DATASETS & lowest sigma (Accuracy) & Middle sigma (Accuracy) & Highest sigma (Accuracy)\\
\hline
1CDT &  0.2 (99.91) & 1 (99.91) & 2 (99.92) \\   
\hline                                                                              
1CHT &  0.2 (99.40) & 1 (99.42) & 2 (99.51) \\       
\hline                                                                              
1CSurr & 1 (91.30) & 1.5 (90.00) & 2 (87.79) \\       
\hline                                                                              
2CDT &  0.2 (58.32) & 0.5 (50.32) & 1 (50.48) \\        
\hline 
2CHT &  0.2 (50.10) & 0.5 (50.89) & 1 (52.15) \\

\hline                                                                              
4CE1CF &  0.2 (97.74) & 0.5 (97.12) & 1.5 (92.40) \\     
\hline                                                                              
4CR &  0.2 (99.99) & 1 (99.99) & 2 (99.99) \\         
\hline                                                                              

4CRE-V2 &  0.2 (20.96) & 0.5 (20.84) & 1 (24.10)  \\    
\hline                                                     
FG\_2C\_2D &  0.2 (95.71) & 0.5 (86.41) & 1 (94.28) \\           
\hline                                                                              
GEARS\_2C\_2D &  0.2 (97.73) & 1 (95.28) & 2 (95.36)  \\       
\hline                                                                              
MG\_2C\_2D &  0.2 (78.03) & 0.5 (78.21) & 1.2 (85.44) \\         
\hline                                                                              
UG\_2C\_2D &  0.2 (70.61) & 0.5 (71.81) & 1 (74.33)  \\   
\hline                                                                UG\_2C\_3D &  0.1 (61.21) & 1 (64.30) & 2 (64.68)  \\   
\hline            
UG\_2C\_5D &  0.5 (77.67) & 1 (80.07) & 1.5 (80.17)  \\      
\hline
keystroke &  0.5 (88.12) & 1 (90.56) & 2 (89.43)  \\      
\hline
\end{tabular}
\vspace{0.3in}
\end{table*}

\section{Conclusion}
This paper provides a comprehensive evaluation of existing approaches that learn from a nonstationary (drifting) environment experiencing extreme verification latency, with respect to classification accuracy, computational complexity and parameter sensitivity. In a nonstationary streaming environment, the nonstationary data, drawn from a drifting distribution, arrive in a streaming manner. The extreme verification latency places an additional constraint that beyond an initial batch, the entire data stream is assumed unlabeled.

The most important contribution of this work is the comprehensive and comparative analysis of the available algorithms in the literature to handle extreme verification latency from three different perspectives: classification accuracy, computational complexity and parameter sensitivity. Our goal in this task has been to determine and describe the relative strengths and weaknesses of these algorithms, and point out different cases and scenarios where one algorithm is better suited over the others. The original COMPOSE algorithm,  COMPOSE with $\alpha$-shape (COMPOSE.V1), was a significant contribution to the field when it was first proposed, as it was the only algorithm capable at the time to address the problem of learning in nonstationary environments in the presence of extreme verification latency with no restrictions on the nature of the data distribution. However, that capability came at a steep price: the algorithm is computationally very expensive (though still significantly more efficient than the Arbitrary Population subTracker (APT) as well as the MClassification). The algorithm also provided to be quite sensitive to the choice of its primary free parameters. Despite these shortcomings, and despite several other competing algorithms developed since then, the original COMPOSE algorithm remains competitive with respect to classification accuracy. The second version of COMPOSE, COMPOSE with GMM (COMPOSE.V2), replaced the $\alpha$-shape based approach for determining the core supports with a Gaussian mixture model based density estimation module that dramatically increased its computational efficiency while retaining the classification accuracy of COMPOSE.V1. The latest version of COMPOSE i.e. FAST COMPOSE, further improves the classification accuracy as well as the computational efficiency compared to all other algorithms. One remaining issue with FAST COMPOSE, however, is its sensitivity to the choice of its primary free parameter, the number of clusters in the cluster-and-label based SSL algorithm used in its core support computation. 

SCARGC was developed as a competing algorithm to the original COMPOSE with the primary advantage of better computational efficiency.  SCARGC with nearest neighbor (1NN) shows comparable accuracy to other algorithms and is less computationally expensive compared to COMPOSE ($\alpha$-shape), and MClassification (but not against COMPOSE with GMM or FAST COMPOSE), while it too is also sensitive to the choice of its primary free parameter -- number of clusters $k$ in k-means clustering based subroutine it uses. SCARGC with SVM while perhaps reasonable with respect to computational burden, was found to be the worst (highest rank) in terms of classification accuracy. SCARGC with SVM retains the high parameter sensitivity as with SCARGC (1NN). 

MClassification shows comparable accuracy performance to other algorithms but appears to be the worst algorithm in computational complexity (other than APT), requiring more runtime than even COMPOSE with $\alpha$-shape on most datasets. This behavior is attributed to its online nature. In fact, MClassification is the only algorithm that is capable of processing the data in an online manner, a distinct advantage in a streaming environment, but that advantage appears to be unrealized or wasted due to the heavy computational burden. This algorithm is also quite sensitive to its primary free parameter.

LEVEL\textsubscript{IW}, as with other algorithms, performed comparably similar with respect to classification accuracy, is less computationally expensive than COMPOSE.V1, SCARGC (1-NN), and MClassification (but more expensive than FAST COMPOSE, SCARGC (SVM) and COMPOSE.V2). While not the best performing algorithm either in terms of classification accuracy or computational efficiency,  LEVEL\textsubscript{IW} has one advantage over other algorithms: greater robustness and stability compared to all of the remaining algorithms with respect to relatively wide fluctuations of the value of its primary free parameter.

\section{Summary of Future Work}
Further work is needed to generate or acquire more challenging datasets, as most algorithms perform similarly on the current synthetic benchmark datasets. Currently, there is a lack of datasets that contain abruptly changing distributions, datasets with recurring concepts or more severe class imbalances, datasets that have substantial feature or class noise, datasets with significant amount of outliers, datasets with very little or almost no shared support, and high dimensional datasets to name a few. 

We already know from the analyses shown in this work that the algorithms described here will not work in all of the above-mentioned scenarios, such as abruptly changing distributions or severe class imbalance. Often in science, however, it is a challenging dataset, or a collection of datasets that provide the motivation for the development of specialized algorithms within a specific disciple. Additionally, future work is needed to provide machine learning community with an algorithm that can perform well with respect to classification accuracy, computationally complexity and parameter sensitivity as well as able to handle challenging datasets mentioned above under initially labeled non-stationary environments.

%

\section*{Acknowledgment}

The authors would like to thank...

\ifCLASSOPTIONcaptionsoff
  \newpage
\fi



%
\bibliographystyle{IEEEtran}
%

\bibliography{bibliography}{}

\begin{thebibliography}{10}
\providecommand{\url}[1]{#1}
\csname url@samestyle\endcsname
\providecommand{\newblock}{\relax}
\providecommand{\bibinfo}[2]{#2}
\providecommand{\BIBentrySTDinterwordspacing}{\spaceskip=0pt\relax}
\providecommand{\BIBentryALTinterwordstretchfactor}{4}
\providecommand{\BIBentryALTinterwordspacing}{\spaceskip=\fontdimen2\font plus
\BIBentryALTinterwordstretchfactor\fontdimen3\font minus
  \fontdimen4\font\relax}
\providecommand{\BIBforeignlanguage}[2]{{%
\expandafter\ifx\csname l@#1\endcsname\relax
\typeout{** WARNING: IEEEtran.bst: No hyphenation pattern has been}%
\typeout{** loaded for the language `#1'. Using the pattern for}%
\typeout{** the default language instead.}%
\else
\language=\csname l@#1\endcsname
\fi
#2}}
\providecommand{\BIBdecl}{\relax}
\BIBdecl

\bibitem{grossberg1988nonlinear}
S.~Grossberg, ``Nonlinear neural networks: Principles, mechanisms, and
  architectures,'' \emph{Neural networks}, vol.~1, no.~1, pp. 17--61, 1988.

\bibitem{muhlbaier2007multiple}
M.~D. Muhlbaier and R.~Polikar, ``Multiple classifiers based incremental
  learning algorithm for learning in nonstationary environments,'' in
  \emph{Machine Learning and Cybernetics, 2007 International Conference on},
  vol.~6.\hskip 1em plus 0.5em minus 0.4em\relax IEEE, 2007, pp. 3618--3623.

\bibitem{karnick2008learning}
M.~Karnick, M.~Ahiskali, M.~D. Muhlbaier, and R.~Polikar, ``Learning concept
  drift in nonstationary environments using an ensemble of classifiers based
  approach,'' in \emph{Neural Networks, 2008. IJCNN 2008.(IEEE World Congress
  on Computational Intelligence). IEEE International Joint Conference
  on}.\hskip 1em plus 0.5em minus 0.4em\relax IEEE, 2008, pp. 3455--3462.

\bibitem{Elwell11}
R.~Elwell and R.~Polikar, ``Incremental learning of concept drift in non-
  stationary environments,'' \emph{IEEE Transactions Neural Networks}, vol.~22,
  no.~10, pp. 1517--1531, 2011.

\bibitem{Ditzler13}
G.~Ditzler and R.~Polikar, ``Incremental learning of concept drift from
  streaming imbalanced data,'' \emph{IEEE Transactions on Knowledge and Data
  Engineering}, vol.~25, pp. 2283--2301, 2013.

\bibitem{Kolter07}
J.~Kolter and M.~Maloof, ``Dynamic weighted majority: An ensemble method for
  drifting concepts,'' \emph{Journal of Machine Learning Research}, vol.~8, pp.
  2755--2790, July 2007.

\bibitem{Street01}
W.~N. Street and Y.~Kim, ``A streaming ensemble algorithm (sea) for large-scale
  classification,'' \emph{ACM SIGKDD international conference on Knowledge
  discovery and data mining}, pp. 377--382, 2001.

\bibitem{Shimodaira00}
H.~Shimodaira, ``Improving predictive inference under covariate shift by
  weighting the log-likelihood function,'' \emph{Journal of statistical
  planning and inference}, vol.~90, no.~2, pp. 227--244, 2000.

\bibitem{Candela09}
J.~Quionero-Candela, M.~Sugiyama, A.~Schwaighofer, and N.~D. Lawrence,
  \emph{Dataset shift in machine learning}.\hskip 1em plus 0.5em minus
  0.4em\relax The MIT Press, 2009.

\bibitem{Ditzler11}
G.~Ditzler and R.~Polikar, ``Semi-supervised learning in nonstationary
  environments,'' \emph{International Joint Conference on Neural Networks}, pp.
  2741--2748, 2011.

\bibitem{Capo13}
K.~D. Robert~Capo and R.~Polikar, ``Active learning in nonstationary
  environments,'' \emph{International Joint Conference on Neural Networks}, pp.
  1--8, 2013.

\bibitem{Marrs10}
G.~Marrs, R.~Hickey, and M.~Black, ``The impact of latency on online
  classiﬁcation learning with concept drift,'' \emph{Knowledge Science,
  Engineering and Management}, vol. 6291.

\bibitem{Dyer14}
K.~B. Dyer, R.~Capo, and R.~Polikar, ``Compose: A semisupervised learning
  framework for initially labeled nonstationary streaming data,'' \emph{IEEE
  Transactions on Neural Networks and Learning Systems}, vol.~25, no.~1, pp.
  12--26, 2014.

\bibitem{Krempl11}
G.~Krempl, ``The algorithm apt to classify in concurrence of latency and
  drift,'' \emph{Intelligent Data Analysis}, pp. 223--233, 2011.

\bibitem{Souza15_SCARGC}
V.~M.~A. Souza, D.~F. Silva, J.~Gama, and G.~E. A. P.~A. Batista, ``Data stream
  classiﬁcation guided by clustering on nonstationary envi- ronments and
  extreme veriﬁcation latency,'' \emph{SIAM International Conference on Data
  Mining}, pp. 873--881, 2015.

\bibitem{Souza15_MClassification}
V.~M.~A. Souza, D.~F. Silva, G.~E. A. P.~A. Batista, and J.~Gama,
  ``Classification of evolving data streams with infinitely delayed labels,''
  \emph{IEEE International Conference on Machine Learning and Applications
  (ICMLA)}, pp. 214--219, 2015.

\bibitem{Zhu02}
X.~Zhu and Z.~Ghahramani, ``Learning from labeled and unlabeled data with label
  propagation,'' Carnegie Mellon University, Tech. Rep. Technical Report
  CMU-CALD-02-107, 2002.

\bibitem{Bennett99}
K.~Bennett and A.~Demiriz, ``Semi-supervised support vector machines,''
  \emph{Advances in Neural Information processing systems}, pp. 368--384, 2002.

\bibitem{Sarnelle15}
J.~Sarnelle, A.~Sanchez, R.~Capo, J.~Haas, and R.~Polikar, ``Quantifying the
  limited and gradual concept drift assumption,'' \emph{International Joint
  Conference on Neural Networks}, 2015.

\bibitem{7849962}
M.~Umer, C.~Frederickson, and R.~Polikar, ``Learning under extreme verification
  latency quickly: Fast compose,'' in \emph{2016 IEEE Symposium Series on
  Computational Intelligence (SSCI)}, Dec 2016, pp. 1--8.

\bibitem{hachiya2012importance}
H.~Hachiya, M.~Sugiyama, and N.~Ueda, ``Importance-weighted least-squares
  probabilistic classifier for covariate shift adaptation with application to
  human activity recognition,'' \emph{Neurocomputing}, vol.~80, pp. 93--101,
  2012.

\bibitem{arostegi2018concept}
M.~Arostegi, A.~I. Torre-Bastida, J.~L. Lobo, M.~N. Bilbao, and J.~Del~Ser,
  ``Concept tracking and adaptation for drifting data streams under extreme
  verification latency,'' in \emph{International Symposium on Intelligent and
  Distributed Computing}.\hskip 1em plus 0.5em minus 0.4em\relax Springer,
  2018, pp. 11--25.

\bibitem{razavi2019novelty}
R.~Razavi-Far, E.~Hallaji, M.~Saif, and G.~Ditzler, ``A novelty detector and
  extreme verification latency model for nonstationary environments,''
  \emph{IEEE Transactions on Industrial Electronics}, vol.~66, no.~1, pp.
  561--570, 2019.

\bibitem{hoffman2014continuous}
J.~Hoffman, T.~Darrell, and K.~Saenko, ``Continuous manifold based adaptation
  for evolving visual domains,'' in \emph{Proceedings of the IEEE Conference on
  Computer Vision and Pattern Recognition}, 2014, pp. 867--874.

\end{thebibliography}

%
%

%

\begin{IEEEbiographynophoto}{Muhammad Umer}
Biography text here.
\end{IEEEbiographynophoto}


\begin{IEEEbiographynophoto}{Robi Polikar}
Biography text here.
\end{IEEEbiographynophoto}




\end{document}